\documentclass{article}

\PassOptionsToPackage{numbers, compress}{natbib}

\usepackage[preprint]{neurips_2023}

\usepackage[utf8]{inputenc} 
\usepackage[T1]{fontenc}    

\usepackage{times}
\usepackage{epsfig}
\usepackage{graphicx}
\usepackage{amsmath}
\usepackage{amssymb}
\usepackage{xcolor}

\usepackage{hyperref}       
\usepackage{url}            
\usepackage{booktabs}       
\usepackage{amsfonts}       
\usepackage{nicefrac}       
\usepackage{microtype}      
\usepackage{xcolor}         
\usepackage{multirow}
\usepackage{xspace}
\usepackage{dsfont}

\newcommand{\myred}[1]{\textcolor{red}{#1}}
\newcommand{\myyellow}[1]{\textcolor{yellow!85!black!}{#1}}
\newcommand*{\eg}{e.g.\@\xspace}
\newcommand*{\ie}{i.e.\@\xspace}
\newcommand*{\etal}{et al.\@\xspace}
\makeatletter
\newcommand*{\etc}{%
    \@ifnextchar{.}%
        {etc}%
        {etc.\@\xspace}%
}
\makeatother

\title{Read, look and detect: Bounding box annotation from image-caption pairs}

\author{%
  Eduardo Hugo Sanchez \\
  IRT Saint Exupéry\\
  Toulouse, France\\
  \texttt{eduardo.sanchez@irt-saintexupery.com} \\
}

\begin{document}

\maketitle

\begin{abstract}
Various methods have been proposed to detect objects while reducing the cost of data annotation. For instance, weakly supervised object detection (WSOD) methods rely only on image-level annotations during training. Unfortunately, data annotation remains expensive since annotators must provide the categories describing the content of each image and labeling is restricted to a fixed set of categories. In this paper, we propose a method to locate and label objects in an image by using a form of weaker supervision: image-caption pairs. By leveraging recent advances in vision-language (VL) models and self-supervised vision transformers (ViTs), our method is able to perform phrase grounding and object detection in a weakly supervised manner. Our experiments demonstrate the effectiveness of our approach by achieving a 47.51\% recall@1 score in phrase grounding on Flickr30k Entities and establishing a new state-of-the-art in object detection by achieving 21.1 $\text{mAP}_{50}$ and 10.5 $\text{mAP}_{50:95}$ on MS COCO when exclusively relying on image-caption pairs. 
\end{abstract}

\section{Introduction}
\label{sec:Intro}

Locating and classifying objects within an image is a fundamental task in computer vision that enables the development of more complex tasks such as image captioning~\cite{yang2017image}, visual reasoning~\cite{GLIP}, among others. Nevertheless, the success of object detection models~\cite{YOLOv5,ren2015faster} typically relies on human supervision in the form of bounding box annotations. In particular, data annotation is a time-consuming and arduous task that requires annotators to draw bounding boxes around objects and label each bounding box with a category from a fixed set of categories. Furthermore, modifying the number of categories may require annotators to relabel or add new bounding boxes.

Several approaches have been proposed to reduce the cost of data annotation in object detection by using image-level labels~\cite{WSDDN,WSCCN,tang2018pcl,W2F,gao2019c,zeng2019wsod2,ren2020instance,huang2020comprehensive}, a dataset containing both labeled and unlabeled data~\cite{STAC,liu2021unbiased,liu2022unbiased} or sparsely-annotated data ~\cite{xu2019missing,zhang2020solving,wang2021comining,li2022siod}. For instance, WSOD methods only use image-level annotations along with the multiple instance learning (MIL)~\cite{maron1997framework} approach. However, the annotation effort is still significant and similar to that required for supervised classification.

In this paper, we take a step forward by learning to locate and label objects within an image from image-caption pairs. Not only captions provide a more natural description of the image content than image-level labels but also constitute a form of weaker supervision since image-caption pairs are easier to collect in vast amounts (\eg from the Web~\cite{sharma2018conceptual}). Our approach combines recent advances in vision-language (VL) models~\cite{ALBEF} and self-supervised vision transformers (ViTs)~\cite{DINO}.

VL models leverage large-scale image-caption datasets and have strong performance on zero-shot image classification, image-text retrieval, and visual reasoning tasks. These models align images with their corresponding captions via contrastive learning. Notably, models that include a cross-modality encoder seem to implicitly learn a more fine-grained word-region alignment without using additional supervision~\cite{ALBEF}. We propose to use the location ability of VL models to automatically annotate objects of interest mentioned in captions. Moreover, VL models do not require retraining when the number of categories to annotate changes as they are already included in the VL model's vocabulary.

Despite the strong ability of VL models to locate objects, they aim at the most distinctive part of the object rather than the whole object. For example, ALBEF~\cite{ALBEF} and VilBERT~\cite{ViLBERT} perform phrase grounding by ranking the object proposals provided by the supervised detector MattNet~\cite{yu2018mattnet}. On the other hand, recent work has shown that representations from self-supervised ViTs contain explicit information about the scene layout of images and produce heatmaps that highlight salient objects~\cite{DINO}. LOST~\cite{LOST} and TokenCut~\cite{wang2022self} show the effectiveness of self-supervised ViT representations to perform unsupervised object discovery and detection without any labels. 

\begin{figure}[t]
\centering
\includegraphics[width=1.0\textwidth]{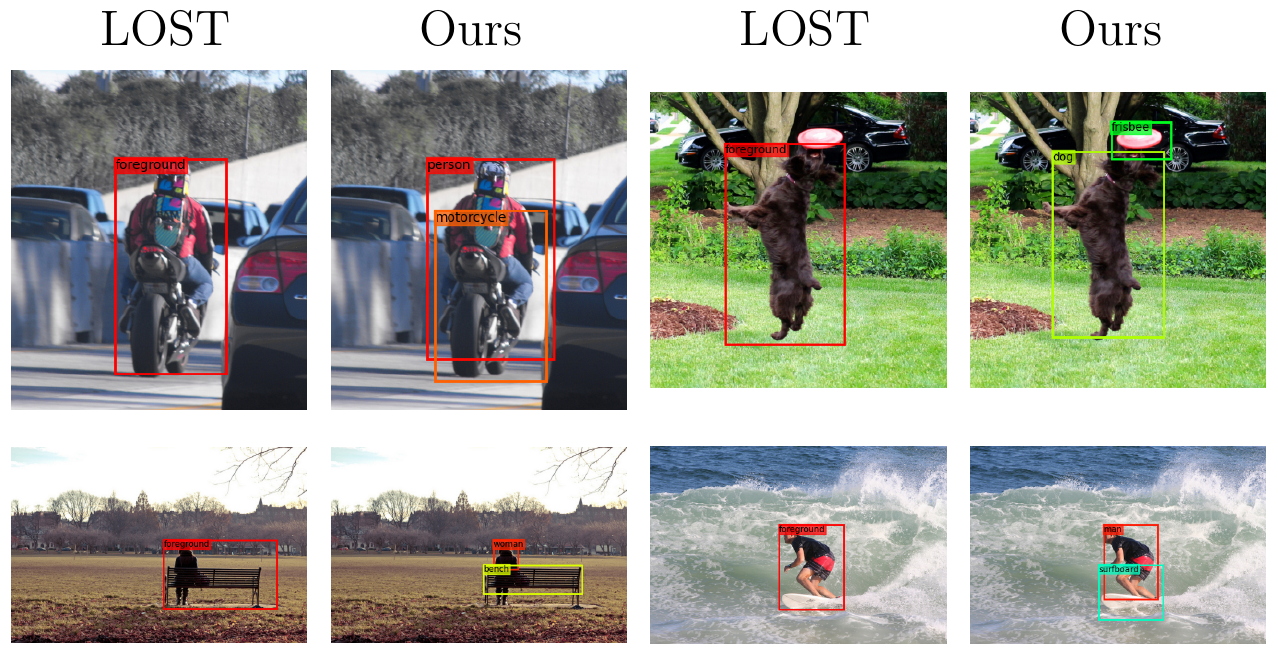}
\caption{Our approach leverages language from captions via ALBEF to annotate multiple objects per image (\eg we detect a dog and a frisbee while LOST generates a single categoryless bounding box).}\label{fig:ImprovementOverLostExamples}
\end{figure}

We make the following contributions in this work. First, we propose a novel method to locate and label objects in images by combining the ability of VL models to point at objects and the ability of self-supervised ViTs to extract whole objects in Section~\ref{sec:Method}. By building upon ALBEF~\cite{ALBEF} and LOST~\cite{LOST}, our method is able to locate multiple objects and generate accurate bounding boxes without human supervision. Figure~\ref{fig:ImprovementOverLostExamples} illustrates the improved ability of our model over LOST. Second, we use our approach to perform phrase grounding and object detection in a weakly supervised fashion. In Section~\ref{sec:Experiments}, we demonstrate that our method achieves competitive performance in phrase grounding on Flickr30k Entities~\cite{Flickr30kEntities} and establish a new state-of-the-art in object detection on MS COCO~\cite{MSCOCO} when exclusively relying on image-caption pairs as unique source of supervision. Additionally, we perform ablation experiments to investigate the key components of our approach, transfer learning experiments on PASCAL VOC2007~\cite{PASCALVOC} and pseudo-labeling experiments to improve the performance in WSOD. In Section~\ref{sec:Conclusion}, we discuss the limitations, future work and conclusions of our work.
\section{Related work}
\label{sec:RW}

\paragraph{Weakly supervised object detection:} To reduce the cost of data annotation, several methods propose to train object detectors using only image-level annotations without the need for bounding box annotations. WSDDN~\cite{WSDDN} introduces the first end-to-end WSOD framework that adopts MIL~\cite{maron1997framework}. Since then, several improvements have been proposed: PCL~\cite{tang2018pcl} performs clustering to improve object proposals and W2F~\cite{W2F} leverages pseudo-label mining from a WSOD model to train a supervised object detector. C-MIDN~\cite{gao2019c} introduces a method for coupling proposals to prevent the detector from capturing the most discriminative object part rather than the whole object. WSOD$^2$~\cite{zeng2019wsod2} performs pseudo-label mining and incorporates a bounding box regressor to fine-tune the location of each proposal. Likely, MIST~\cite{ren2020instance} performs pseudo-label mining where highly overlapping proposals are assigned to the same label. CASD~\cite{huang2020comprehensive} combines self-distillation with multiple proposal attention maps generated via data augmentation. Closely related to our work, Cap2Det~\cite{CAP2DET} learns from image-caption pairs by extracting image-level annotations from captions using a supervised text classifier. These predicted image-level annotations are subsequently used to train a WSOD model based on MIL. Additionally, Cap2Det~\cite{CAP2DET} refines the WSOD model by retraining on instance-level pseudo-labels multiple times. Most of the existing WSOD methods rely on object proposal algorithms (e.g. Selective Search \cite{uijlings2013selective} or Edge Boxes\cite{zitnick2014edge}). By exclusively leveraging self-supervision on image-caption pairs, our approach outperforms the state-of-the-art model Cap2Det~\cite{CAP2DET} without the need for a supervised text classifier. Additionally, our approach outperforms relevant WSOD baselines~\cite{tang2018pcl,gao2019c} that use a form of stronger supervision (image-level annotations) and object proposal algorithms.
\paragraph{Learning from unlabeled or partially labeled data:} Some approaches alleviate the lack of bounding box annotations by leveraging a small labeled dataset and a large unlabeled dataset via semi-supervised learning~\cite{STAC,liu2021unbiased, liu2022unbiased} and active learning~\cite{wang2018towards,vo2022active}. Li \etal~\cite{li2022siod} propose to train an object detector using only a single instance annotation per category per image. Other methods combine image-level and instance-level pseudo-annotations during training~\cite{xu2019missing,ren2020instance}. Sohn \etal~\cite{STAC} propose a two-stage training in which an object detector is trained on available labeled data. This model is subsequently used to select high-confidence bounding boxes on unlabeled data as pseudo-labels. Wang \etal~\cite{wang2021comining} address the missing annotation problem by introducing a siamese network where each branch is used to generate pseudo-labels for each other. Likely, recent work~\cite{liu2021unbiased,liu2022unbiased} leverage the teacher-student framework in object detection. In this work, we also explore the use of pseudo-labels to improve WSOD performance.
\paragraph{VL models:} Learning joint VL representations from image-caption pairs in a self-supervised fashion has proven to be effective to perform multiple downstream tasks~\cite{ViLBERT,ViLBERT2} such as visual question answering, image retrieval, image captioning, zero-shot classification, \etc. VL models~\cite{LXMERT,VisualBERT,VLBERT,VILT,VILLA,OSCAR,UNITER,ALIGN,CLIP,ALBEF} are generally trained on a combination of loss functions: masked language modelling (MLM), where a masked word token is predicted; masked image modelling (MIM), where a masked image region feature or object category is predicted, image-text contrastive learning (ITC), where positive/negative image-caption pairs are assigned to high/low similarity scores, respectively; and image-text matching (ITM), that predicts whether an image and a caption match. Many strategies have been proposed to achieve improved VL representations.  VisualBERT~\cite{VisualBERT} uses a supervised object detector to extract visual embeddings. VILLA~\cite{VILLA} performs adversarial training in the representations space. OSCAR~\cite{OSCAR} uses object tags to ease VL alignment. UNITER~\cite{UNITER} encourages alignment between words and image regions extracted by an object detector. More recently, CLIP~\cite{CLIP} leverages a massive amount of image-caption pairs and achieves impressive performance at zero-shot classification. However, CLIP underperforms at other VL tasks as the interaction between vision and language is very shallow (\ie a simple dot product). Li \etal~\cite{ALBEF} propose a new model called ALBEF, which builds upon previous models~\cite{LXMERT,ViLBERT,CLIP,VILT,ALIGN} and is composed of a vision encoder, a language encoder, and a cross-modality encoder for deeper VL interaction. By leveraging a large image-caption dataset~\cite{sharma2018conceptual}, ALBEF outperforms previous models at many VL tasks without the need for a supervised object detector to extract region-based image representations. Our approach leverages a pre-trained ALBEF model to locate the image region that corresponds to a word or a textual description.
\paragraph{Open vocabulary detection:} Classifying an object or image has been traditionally limited to a small set of fixed categories. Zhang \etal~\cite{zhang2016online} leverages the vocabulary from image-caption datasets to perform image classification across more than 30k classes. The recent success of VL models~\cite{CLIP,ALBEF} has motivated other methods to leverage image-caption pairs to perform object detection on a larger number of categories. Zareian \etal~\cite{zareian2021open} use bounding box annotations from base classes to perform correctly in target classes mentioned in captions. Gao \etal~\cite{gao2022open} use a supervised object detector trained on MS COCO~\cite{MSCOCO} to generate pseudo-bounding box annotations for categories mentioned in captions. Similar approaches~\cite{zhong2022regionclip,shi2022proposalclip} have been proposed by extending CLIP~\cite{CLIP}. We also leverage VL models to annotate objects using an arbitrary number of categories in a self-supervised manner without relying on bounding box annotations like previous methods.
\paragraph{Object discovery:} Recently, several studies explore methods for object localization that rely solely on visual cues. LOST~\cite{LOST} extracts image representations via a self-supervised ViT~\cite{DINO} which are subsequently used to identify the image patches corresponding to an object based on their correlation. Wang \etal~\cite{wang2022self} also leverage DINO representations which are used to build a graph. A normalized graph-cut is used to split the foreground object from the background. Both methods can only locate a single object per image without providing its category. Our approach builds upon LOST by integrating the language modality, enabling it to locate and label multiple objects per image.
\begin{figure}
\includegraphics[width=1.0\textwidth]{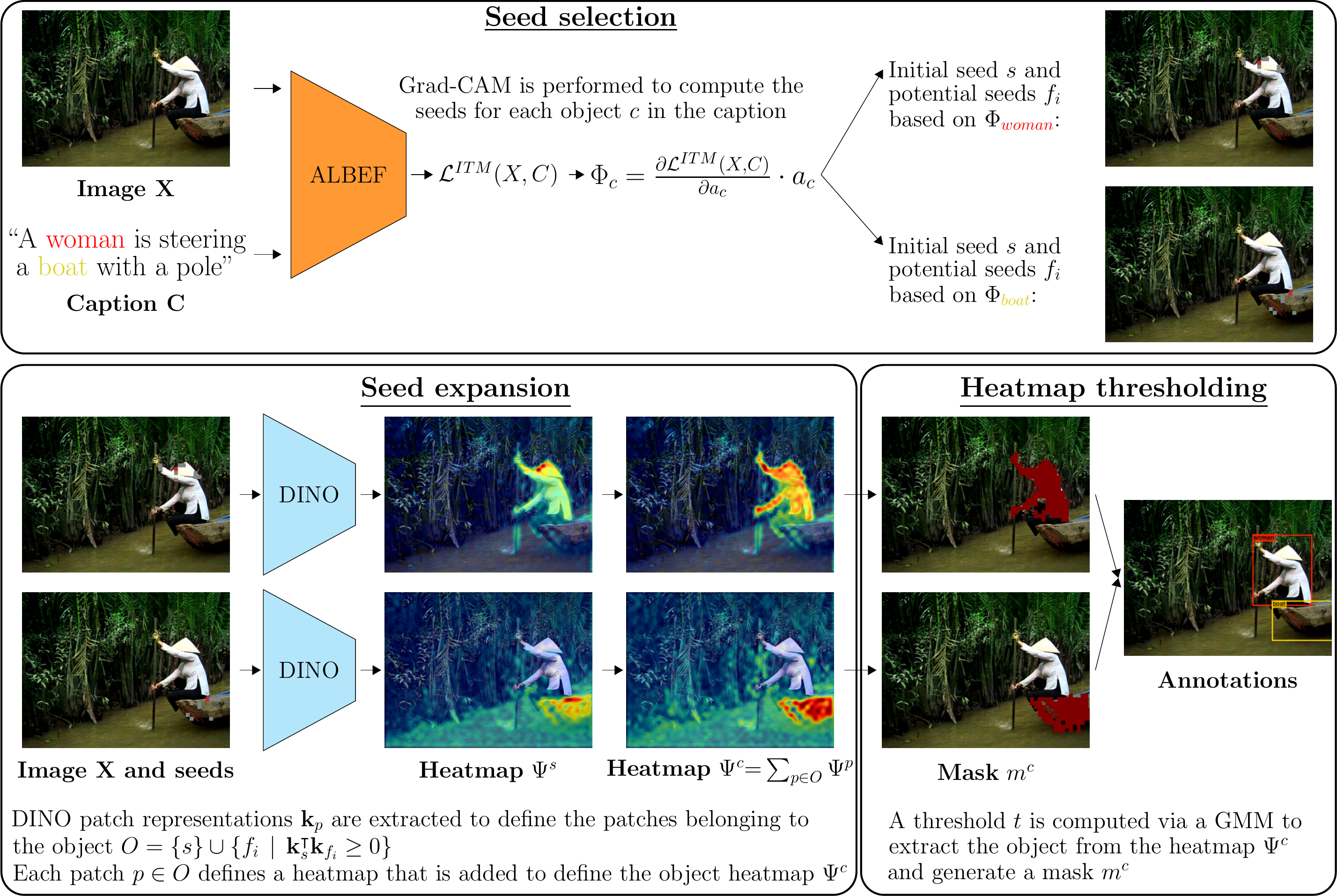}
\caption{Bounding box generation using the caption "A \myred{woman} is steering a \myyellow{boat} with a pole". First, we select the \textit{initial} and \textit{potential seeds} (red and gray patches, respectively) via a VL model for each category identified in the caption. Second, we perform \textit{seed expansion} by measuring similarity between patches via a ViT. Finally, each heatmap is thresholded and a bounding box is drawn on top.}
\label{fig:Overview_plus_plus}
\end{figure}

\section{Method}
\label{sec:Method}

To annotate objects from image-caption pairs, our approach consists of two main stages. First, we leverage the cross-modality encoder from a pre-trained VL model to automatically select the image patches (or \textit{seeds}) that may belong to a given object (defined by a word token or a set of word tokens). The \textit{seed selection} process is described in Section~\ref{sec:Pointing}. Second, we use a self-supervised ViT to compute the similarity between image patches. Intra-image similarity is used to filter out image patches selected in the first stage and generate a heatmap corresponding to the object. This process is known as \textit{seed expansion}. Then, a heatmap threshold is computed via a Gaussian mixture model (GMM) to separate the object patches from the background ones. Finally, a bounding box enclosing the object patches is generated. Section~\ref{sec:Extending} describes the process to generate a heatmap and extract an object from it. Figure \ref{fig:Overview_plus_plus} shows an overview of our approach.

\subsection{Pointing at objects with VL models}
\label{sec:Pointing}

Our proposed method is motivated by the observation that VL models implicitly learn to align words in the captions with patches in the images even though these models are only trained to align images with their corresponding captions~\cite{ALBEF}. Furthermore, we can annotate a large amount of objects since the number of objects categories is as large as the vocabulary used in the captions during training of VL models. We leverage the ability of VL models to point at objects and the fact that most of the salient objects in an image are mentioned in its respective caption~\cite{OSCAR}.

In this section, we explain how the fine-grained alignment between words and patches is computed in VL models implementing a cross-modality encoder (\eg ALBEF~\cite{ALBEF}) and how we leverage it to point at objects in an image. Let $X = \{ p_1, p_2, \dots, p_{N_P} \}$ be an image composed of $N_P$ patches and $C = \{ w_1, w_2, \dots, w_{N_T} \}$ be its corresponding caption composed of $N_T$ word tokens. An image encoder and text encoder are used to extract image and text representations which are both fed into the cross-modality encoder. In the $l^{vl}$-th cross-attention layer of this encoder, we compute the value and key representations for each image patch, \ie $V = \{ \mathbf{v}_0, \mathbf{v}_1, \dots, \mathbf{v}_{N_P} \}$ and $K = \{ \mathbf{k}_0, \mathbf{k}_1, \dots, \mathbf{k}_{N_P} \}$, respectively, where $\mathbf{v}_0$ and $\mathbf{k}_0$ are the representations of the classification token [CLS]. 

Given a word token of interest $w_c$ (\eg `person', `dog', \etc.), we compute its query representation $\mathbf{q}_{c}$. The relation between the word token $w_c$ and the image patches $\{p_i\}_{i=1}^{N_P}$ is given by the hidden representation $\mathbf{h}_{c}$ as shown in Equation~\ref{eq:CrossAttentionWeights} where $d$ is the dimension of the query representations.
\begin{equation}
\mathbf{h}_{c} = \sum_{i=0}^{N_P} a_{c,i} \cdot \mathbf{v}_{i} \quad \ \text{where} \ a_{c,i} = \frac{\exp(\mathbf{q}_{c}^{\intercal} \mathbf{k}_{i}/\sqrt{d})}{\sum_{j=0}^{N_P} \exp(\mathbf{q}_{c}^{\intercal} \mathbf{k}_{j}/\sqrt{d})}
\label{eq:CrossAttentionWeights}
\end{equation}
As observed, the hidden representation of $w_c$ is a linear combination of the value representations corresponding to the image patches. Furthermore, these representations are weighted according to the attention scores $a_{c,i}$ that implicitly provide the similarity between $w_c$ and $p_i$ via the product $\mathbf{q}_{c}^{\intercal} \mathbf{k}_{i}$. Through the use of the cross-modality encoder, one can identify the image regions that are most closely aligned with a particular word token. We use Grad-CAM~\cite{selvaraju2017grad} to rank the image patches in order of importance. Equation~\ref{eq:GradCAMScore} displays the importance score of the image patch $p_i$ with respect to the word token $w_c$ where $\mathcal{L}^{ITM}(X,C)$ is the binary cross-entropy loss that measures whether the image $X$ and the caption $C$ match or not. When ranking image patches, we do not take into account the attention score corresponding to the classification token [CLS], $a_{c,0}$.
\begin{equation}
\Phi_{c,i} = \frac{\partial \mathcal{L}^{ITM}(X,C)}{\partial a_{c,i}} \cdot a_{c,i}
\label{eq:GradCAMScore}
\end{equation}
Unfortunately, Grad-CAM scores are insufficient to generate an accurate bounding box by themselves (see Section~\ref{sec:Experiments}). For example, Gao \etal~\cite{gao2022open} use a supervised Mask R-CNN~\cite{he2017mask} to generate bounding boxes that cover the activated image patches by the word token $w_c$ for object detection. Similarly, Li \etal~\cite{ALBEF} rank MattNet~\cite{yu2018mattnet} proposals based on Grad-CAM maps for phrase grounding.

However, we observe that while Grad-CAM scores do not highlight the image patches corresponding to the whole object, they are useful to point at the most discriminative parts of it. Therefore, we propose to use a set $D=\{f_i\}_{i=1}^{M}$ of $M$ image patches with the highest score $\Phi_{c,i}$ for a given word token of interest $w_c$ to point at the object. The image patches in $D$ are referred to as \textit{potential seeds} and this process is referred to as \textit{seed selection}. Pointing is a natural way for humans to refer to an object~\cite{bearman2016s} and constitutes the first stage of our proposed approach.

\subsection{Extracting objects with self-supervised ViTs}
\label{sec:Extending}

We make use of the self-supervised ViT capability~\cite{DINO} to measure the similarity between image patches. Using the location information provided in the previous stage, our approach takes advantage of the fact that object patches correlates positively with each other but negatively with background patches. This idea is successfully applied in LOST~\cite{LOST} to perform object discovery. Our work is inspired by LOST and extends its capabilities by incorporating the language modality.

Assuming that the object area is smaller than the background area, LOST uses the patch with the smallest number of positive correlations with other patches in order to point at an object. However, this assumption may not always hold in practice (\eg an object covering more area than the background, multiple objects, \etc.). Compared to LOST, our method is able to generate multiple bounding boxes per image (as many objects as mentioned in the caption). Furthermore, our method can annotate each object with a label while LOST can only retrieve a single object without specifying its category. Figure~\ref{fig:ImprovementOverLostExamples} displays the differences between our approach and LOST.

In this work, we average the first $N$ patch locations with the highest value of $\Phi_{c,i}$ in $D$ to compute the \textit{initial seed} $s$ for a given $w_c$. Following LOST, we extract the key representations of the initial and potential seeds, \ie $\mathbf{k}_{s}$ and $\{\mathbf{k}_{f_i}\}_{i=1}^{M}$, respectively, from the $l^{vit}$-th self-attention layer of a self-supervised ViT~\cite{DINO}. Then, the similarity between the initial seed and potential seeds is computed via the dot product of their respective representations to determine the image patches belonging to the object. We assume that potential seeds that are positively correlated to the initial seed belong to the object while potential seeds that are negatively correlated to the initial seed belong to the background. 

Thus, patches belonging to the object are defined by the set $O = \{s\} \cup \{f_i\mid f_i \in D\ \text{and}\ \mathbf{k}_{s}^{\intercal} \mathbf{k}_{f_i} \geq 0\}$. Each patch $p \in O$ generates a heatmap $\Psi^{p} \in \mathbb{R}^{N_P}$, where the i-th dimension $\Psi^{p}_{i}$ is computed via the dot product between its key representation $\mathbf{k}_{p}$ and the key representation of the patch $p_i$ (also extracted by the ViT), \ie $\mathbf{k}_{p_i}\ \forall i \in \{1,\dots,N_P\}$ as shown in Equation~\ref{eq:GradCAMHeatmap}.
\begin{equation}
\Psi^{p}_{i} = \mathbf{k}_{p}^{\intercal} \mathbf{k}_{p_i}
\label{eq:GradCAMHeatmap}
\end{equation}
Finally, the heatmap of the object $w_c$ is defined by the sum of the heatmaps corresponding to the patches in $O$ as shown in Equation~\ref{eq:GradCAMHeatmapAll}. This process is referred to as \textit{seed expansion}.
\begin{equation}
\Psi^{c}  = \sum_{p \in O} \Psi^{p}
\label{eq:GradCAMHeatmapAll}
\end{equation}
To extract the object from the heatmap $\Psi^{c}$, we define a threshold $t$. While LOST sets $t{=}0$, we assume that patches belonging to the object and background are defined by two normal distributions $p_o{=}\mathcal{N}(\mu_o,\,\sigma^{2}_o)$ and $p_b{=}\mathcal{N}(\mu_b,\,\sigma^{2}_b)$, respectively. The parameters $\mu_o,\ \sigma_o,\ \mu_b,\ \sigma_b \in \mathbb{R}$ are estimated via a GMM per heatmap with $k{=}2$ components. Then, the threshold is calculated by solving $p_o(t){=}p_b(t)$ such that $\mu_b < t < \mu_o$. For small objects, $p_o$ is barely noticeable and hard to estimate via GMM since only the background component is recognizable. We assume only one component is distinguishable when the overlapping between the estimated distributions $p_o$ and $p_b$ is significant (\ie $ \mu_b + 1.5\sigma_b < \mu_o - 1.5\sigma_o$). In such a case, we use the threshold $t=\mu + \gamma\sigma$ where $\gamma$ is a constant and $\mu$ and $\sigma$ are the mean and the standard deviation of $\Psi^{c}$, respectively. Supplementary material provides bounding box examples using multiple $t$ values. To generate a bounding box, a mask $m^{c}$ is obtained by thresholding the heatmap $\Psi^{c}$ as shown Equation~\ref{eq:GradCAMMask} where $\Psi^{c}_{i}$ is the i-th dimension of the heatmap $\Psi^{c}$. Later, a bounding box is drawn by enclosing the segment that includes the initial seed $s$.
\begin{equation}
m^{c}_i = \mathds{1}_{\Psi^{c}_{i} \geq t}
\label{eq:GradCAMMask}
\end{equation}

\section{Experiments and results}
\label{sec:Experiments}

\subsection{Setup details}

\paragraph{Tasks and datasets:} We perform weakly supervised phrase grounding and object detection to demonstrate the effectiveness of our method to annotate objects. In Section~\ref{sec:WSPG}, we present our experimental results for phrase grounding on Flickr30k Entities~\cite{Flickr30kEntities}, an extension of Flickr30k~\cite{young2014image} which consists of $\approx$ 32k images collected from Flickr each of which is described with 5 captions. Image-caption samples are split into $\approx$ 30k training, 1k validation, and 1k test samples. Flickr30k Entities includes manually-annotated bounding boxes that are linked with entities mentioned in captions. Results are reported in terms of recall@1 on the test set. In Section \ref{sec:WSODExp}, we perform WSOD on MS COCO~\cite{MSCOCO} which contains 113k training and 5k validation images. Each image is described with 5 captions. Additionally, the dataset provides bounding box annotations covering 80 object categories such as person, bicycle, car, plane, \etc. We also conduct transfer learning experiments using samples from MS COCO to train an object detector that predicts PASCAL VOC2007~\cite{PASCALVOC} categories since this dataset does not provide captions. PASCAL VOC2007 is an object detection dataset that contains 2501 training, 2510 validation, and 4952 test images. Objects are labeled into 20 classes (\eg person, bird, cat, cow, dog, \etc.). Results are reported in terms of mean average precision at $\text{IoU}{=}0.5$, \ie $\text{mAP}_{50}$, and average mAP over multiple $\text{IoU}$ values ranging from 0.5 to 0.95 with a step of 0.05, \ie $\text{mAP}_{50:95}$. Results are reported on the MS COCO validation set and the PASCAL VOC2007 test set. In all cases, bounding box annotations are only used during evaluation.

\noindent \textbf{Model architecture}: To point at objects, we use ALBEF pre-trained on 14M image-caption pairs~\cite{ALBEF} and fine-tuned on 20k image-caption pairs~\cite{yu2016modeling}. It is worth mentioning that any VL model that includes a cross-modality encoder can be used. To perform seed expansion, we use the self-supervised ViT from DINO (\ie ViT-S/16~\cite{DINO}). For comparative purposes, we also use the image encoder from ALBEF (\ie ViT-B/16~\cite{dosovitskiy2021an}) to compute the similarity between image patches. In WSOD, our approach generates bounding box annotations to train a YOLOv5 object detector~\cite{YOLOv5} in a supervised manner.

\noindent \textbf{Hyperparameters}: We set the VL cross-attention layer to $l^{vl}{=}8$ and the ViT self-attention layer to $l^{vit}{=}11$. To compute the initial seed, we average the first $N{=}3$ patch locations from $D$ and set the number of potential seeds to $M{=}10$. To compute the threshold, we use $\gamma=1.75$. Our experiments are executed on a NVIDIA GeForce RTX 3090. 

\subsection{Weakly supervised phrase grounding}
\label{sec:WSPG}

We conduct experiments on Flickr30k Entities to evaluate the ability of our approach to associate phrases describing objects to image regions. While a single word can define the category of an object, a phrase provides additional attributes (\eg color, size, position, \etc.). Our method processes phrases by simply adding up the heatmaps of each word $w_{c_i}$ in the phrase $P$, \ie $\Psi^{\text{phrase}} = \sum_{c_i \in P} \Psi^{c_i}$.

In Table~\ref{tab:VisualGroundingFlickr30k}, we report our results in terms of recall@1 that represents the ratio of the number of phrases whose ground truth bounding boxes have significant overlap with the generated bounding boxes by our model (\ie $\text{IoU} \geq 0.5$) to the total number of phrases. 

Our baseline model (referred to as ALBEF C-A maps) uses the cross-modality encoder to produce heatmaps $\Phi^{\text{phrase}}$ , which are then thresholded to generate bounding boxes. As shown in Section~\ref{sec:Pointing}, our approach builds upon $\Phi^{\text{phrase}}$ via a self-supervised ViT to generate the expanded heatmaps $\Psi^{\text{phrase}}$. We evaluate two variants of our approach by using the ViT from ALBEF and DINO to generate the object heatmaps (referred to as ALBEF ViT maps and DINO ViT maps, respectively).

\begin{figure}[t]
\centering
\includegraphics[width=1.00\textwidth]{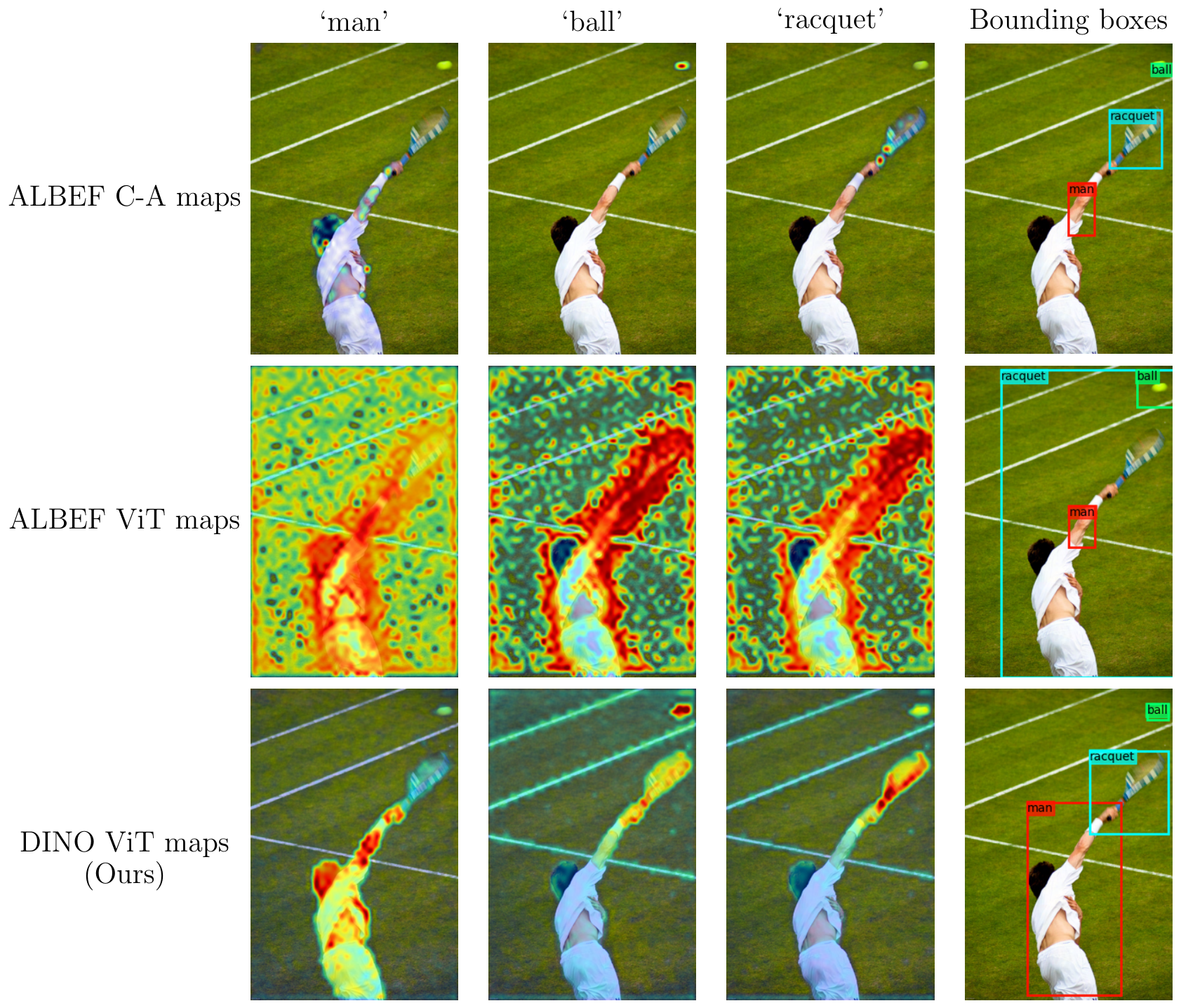}
\caption{Heatmaps and generated bounding boxes corresponding to `man', `ball' and `racquet'. ALBEF C-A maps point successfully at objects while struggle to get the object extent. ALBEF ViT maps tend to be noisier than DINO ViT maps which generate high-quality bounding boxes.}
\label{fig:CompareMethods}
\end{figure}

As observed, the variants ALBEF ViT maps and DINO ViT maps achieve higher performance compared to the baseline (improvements of 7.11\% and 10.65\%, respectively). As hypothesized, the baseline model exhibits limitations in accurately capturing the spatial extent of objects despite its ability to point at them in the image as shown in Figure~\ref{fig:CompareMethods}. Moreover, DINO ViT maps outperform ALBEF ViT maps by a margin of 3.54\%. This difference suggests that DINO's loss function is more effective to capture the underlying relationships between image patches.

For the sake of comparison, we also report the performance of the state-of-the art model for weakly supervised phrase grounding, \ie InfoGround~\cite{gupta2020contrastive}. Our approach achieves a competitive score of 47.51\% comparable to InfoGround performance (47.88\% and 51.67\% when trained on Flickr30k Entities and MS COCO, respectively). Nevertheless, InfoGround uses a Faster R-CNN~\cite{ren2015faster} pre-trained on Visual Genome~\cite{krishna2017visual} to generate object proposals and extract object features. Thus, our approach offers an efficient solution for phrase grounding without the need for an object detector. Our approach represents a promising alternative to InfoGround, particularly in scenarios where the object detector does not include some categories or where obtaining bounding box annotations is difficult.

\begin{table}
\caption{Weakly supervised phrase grounding performance on Flickr30k Entities.}
\centering
\begin{tabular}{llcc}
\toprule
\multirow{2}{*}{Method} & \multirow{2}{*}{Training data} & Supervised object   & \multirow{2}{*}{Recall@1} \\
                        &                                & proposal generator? &  \\
\midrule
ALBEF C-A maps                             & 14M image-caption pairs~\cite{ALBEF}         & No                  & 36.86 \\
ALBEF ViT maps                             & 14M image-caption pairs~\cite{ALBEF}         & No                  & 43.97 \\
\multirow{2}{*}{\textbf{DINO ViT maps}}    & 14M image-caption pairs~\cite{ALBEF}         & \multirow{2}{*}{No} & \multirow{2}{*}{\textbf{47.51}} \\
                                           & + ImageNet images~\cite{DINO}                &                     &  \\
\midrule
InfoGround~\cite{gupta2020contrastive}     & Flickr30k Entities~\cite{Flickr30kEntities}  & Yes, Faster R-CNN~\cite{ren2015faster} & 47.88 \\
InfoGround~\cite{gupta2020contrastive}     & MS COCO~\cite{MSCOCO}                        & Yes, Faster R-CNN~\cite{ren2015faster} & 51.67 \\
\bottomrule
\end{tabular}
\label{tab:VisualGroundingFlickr30k}
\end{table}

\subsection{Weakly supervised object detection}
\label{sec:WSODExp}
We investigate the ability of our approach to perform WSOD. Our methodology involves defining a set of object categories and searching through captions to identify if any of these categories are mentioned. If a category is found, our approach generates a corresponding bounding box as described in Section~\ref{sec:Method}. Then, we train an object detector (\ie Yolov5~\cite{YOLOv5}) from scratch in a supervised manner using the generated bounding box annotations. We evaluate our approach on MS COCO~\cite{MSCOCO} and PASCAL VOC 2012~\cite{PASCALVOC}. While our method is capable of labeling a large number of object categories, we use these datasets as they provide bounding box annotations for evaluation purposes.



\paragraph{Comparison with WSOD methods:} We compare our approach with state-of-the-art WSOD methods to demonstrate its effectiveness in Table~\ref{tab:coco_experiments_state_of_the_art}. Our approach achieves 21.1 $\text{mAP}_{50}$ and 10.5 $\text{mAP}_{50:95}$ on MS COCO outperforming the variants of Cap2Det~\cite{CAP2DET} that learn from image-caption pairs: Cap2Det\textsuperscript{EM} that generates image-level annotations from captions via lexical matching and Cap2Det\textsuperscript{CLSF} that employs a supervised text classifier to process captions and extract image-level annotations. Our approach demonstrates better performance without the need for an object proposal algorithm, a supervised text classifier or using refinement. Compared to methods that learn from image-level annotations~\cite{gao2019c,zeng2019wsod2,ren2020instance,huang2020comprehensive}, our approach demonstrates competitive performance and outperforms relevant baselines such as PCL~\cite{tang2018pcl} and C-MIDN~\cite{gao2019c} (8.5 $\text{mAP}_{50:95}$ and 9.6 $\text{mAP}_{50:95}$, respectively) by achieving 10.5 $\text{mAP}_{50:95}$. It is worth noting that these WSOD methods rely on pseudo-labeling techniques and image-level annotations that constitute a form of stronger supervision. For the sake of comparison, we also report the results of Yolov5 trained in a fully-supervised manner.

\paragraph{Transfer learning and pseudo-labeling (P-L):} Due to the lack of captions in PASCAL VOC2007, our approach generates annotations by searching PASCAL VOC2007 object categories from MS COCO image-caption pairs. Results in terms of $\text{mAP}_{50}$ per category are reported in Table~\ref{tab:pascal_experiments_state_of_the_art} where best results are highlighted in bold. Our approach achieves 40.9 $\text{mAP}_{50}$ outperforming Cap2Det\textsuperscript{EM} (39.9 $\text{mAP}_{50})$ while being behind Cap2Det\textsuperscript{CLSF} (43.1 $\text{mAP}_{50}$). To further improve performance, we propose a simple pseudo-labeling (P-L) technique. First, we use the trained object detector to generate predictions on the training images of PASCAL VOC2007. Pseudo-labels are selected by setting the confidence and IoU thresholds to 0.2 and 0.5, respectively in the NMS algorithm. Then, we fine-tune our trained object detector on these pseudo-labels. We report an improvement of 1.6 $\text{mAP}_{50}$ and 1.1 $\text{mAP}_{50:95}$. Despite the global $\text{mAP}_{50}$ being inferior to that of Cap2Det\textsuperscript{CLSF}, it is worth noting that our approach implementing P-L outperforms Cap2Det\textsuperscript{CLSF} in many categories.

\begin{table}
\caption{Comparison with WSOD models on MS COCO.}
\centering
\begin{tabular}{lllcc}
\toprule
Model & Supervision source & $\text{mAP}_{50}$ & $\text{mAP}_{50:95}$ \\
\midrule
Cap2Det\textsuperscript{EM}~\cite{CAP2DET}       & image-captions pairs & 19.7 & 8.9 \\ 
Cap2Det\textsuperscript{CLSF}~\cite{CAP2DET} & image-captions pairs & 20.2 & 9.1 \\ 
\textbf{Ours}               & image-captions pairs & \textbf{21.1} & \textbf{10.5} \\ 
\midrule
PCL~\cite{tang2018pcl}              & image-level annotations & 19.4 & 8.5 \\ 
C-MIDN~\cite{gao2019c}              & image-level annotations & 21.4 & 9.6 \\ 
WSOD$^2$~\cite{zeng2019wsod2}       & image-level annotations & 22.7 & 10.8 \\ 
MIST~\cite{ren2020instance}         & image-level annotations & 25.8 & 12.4 \\ 
CASD~\cite{huang2020comprehensive}  & image-level annotations & 26.4 & 12.8 \\ 
\midrule
Fully supervised~\cite{YOLOv5}      & bounding box annotations & 66.2 & 46.7 \\ 
\bottomrule
\end{tabular}
\label{tab:coco_experiments_state_of_the_art}
\end{table}

\paragraph{Ablation experiments:} We  also perform ablation experiments to identify the key components of our approach in WSOD. To annotate objects, we employ the variants of our approach presented in Section~\ref{sec:WSPG}, \ie ALBEF C-A maps, ALBEF ViT maps and DINO ViT maps. Tables~\ref{tab:coco_experiments_ablation} and~\ref{tab:pascal_experiments_ablation} display the results of our experiments on MS COCO and PASCAL VOC2007, respectively. As observed, ALBEF C-A maps perform poorly at object detection achieving the lowest scores $\text{mAP}_{50}$ and $\text{mAP}_{50:95}$. While ALBEF C-A maps are able to accurately point at objects, they fail to correctly detect their extent. On the other hand, self-supervised ViTs (ALBEF ViT maps and DINO ViT maps) are effective to capture the extent of objects through the seed expansion. In MS COCO, DINO ViT maps outperform ALBEF ViT maps as expected since DINO ViT maps are less noisy and generates visually more accurate bounding boxes as shown in Figure~\ref{fig:CompareMethods}. Surprisingly, ALBEF ViT maps achieve slightly better results than DINO ViT maps in PASCAL VOC2007.

\begin{table}[!h]
\caption{Comparison with WSOD models on PASCAL VOC2007.}
\centering
\resizebox{\textwidth}{!}{\begin{tabular}{lccccccccccccccccccccc}
\toprule
Model & \rotatebox[origin=c]{90}{aero} & \rotatebox[origin=c]{90}{bike} & \rotatebox[origin=c]{90}{bird} & \rotatebox[origin=c]{90}{boat} & \rotatebox[origin=c]{90}{bottle} & \rotatebox[origin=c]{90}{bus} & \rotatebox[origin=c]{90}{car} & \rotatebox[origin=c]{90}{cat} & \rotatebox[origin=c]{90}{chair} & \rotatebox[origin=c]{90}{cow} & \rotatebox[origin=c]{90}{table} & \rotatebox[origin=c]{90}{dog} & \rotatebox[origin=c]{90}{horse} & \rotatebox[origin=c]{90}{mbike} & \rotatebox[origin=c]{90}{person} & \rotatebox[origin=c]{90}{plant} & \rotatebox[origin=c]{90}{sheep} & \rotatebox[origin=c]{90}{sofa} & \rotatebox[origin=c]{90}{train} & \rotatebox[origin=c]{90}{tv} & $\text{mAP}_{50}$\\
\midrule
Cap2Det\textsuperscript{EM}~\cite{CAP2DET}       & 63.0 & 50.3 & 50.7 & 25.9 & \textbf{14.1} & 64.5 & \textbf{50.8} & 33.4 & \textbf{17.2} & 49.0 & 48.2 & 46.7 & 44.2 & 59.2 & 10.4 & 14.3 & 49.8 & 37.7 & 21.5 & 47.6 & 39.9 \\
Cap2Det\textsuperscript{CLSF}~\cite{CAP2DET}     & \textbf{63.8} & 42.6 & 50.4 & 29.9 & 12.1 & 61.2 & 46.1 & 41.6 & 16.6 & \textbf{61.2} & \textbf{48.3} & \textbf{55.1} & \textbf{51.5} & 59.7 & 16.9 & 15.2 & 50.5 & \textbf{53.2} & 38.2 & \textbf{48.2} & \textbf{43.1} \\
Ours                                             & 58.8 & 64.6 & 52.3 & 28.9 & 10.0 & 57.2 & 42.2 & \textbf{50.7} & 12.8 & 54.3 & 32.4 & 38.8 & 37.4 & \textbf{61.9} & \textbf{24.2} & 17.6 & 47.3 & 39.0 & 52.3 & 34.4 & 40.9 \\
Ours + P-L                                       & 56.1 & \textbf{68.5} & \textbf{55.6} & \textbf{31.1} & 12.3 & \textbf{64.8} & 48.6 & 48.8 & 15.5 & 57.8 & 22.9 & 34.8 & 42.3 & 59.1 & 23.2 & \textbf{19.1} & \textbf{51.8} & 42.8 & \textbf{54.8} & 41.0 & 42.5\\
\midrule
Supervised~\cite{YOLOv5}                   & 70.2 & 74.3 & 42.8 & 40.4 & 40.8 & 73.6 & 83.3 & 62.0 & 37.7 & 61.3 & 58.3 & 56.1 & 77.5 & 71.2 & 78.0 & 35.5 & 50.5 & 55.0 & 75.1 & 60.2 & 60.2 \\
\bottomrule
\end{tabular}}
\label{tab:pascal_experiments_state_of_the_art}
\end{table}

\begin{table}[!h]
\centering
\begin{minipage}[t]{0.48\linewidth}\centering
\caption{Ablation experiments on MS COCO.}
\begin{tabular}{llcc}
\toprule
Method & $\text{mAP}_{50}$ & $\text{mAP}_{50:95}$ \\
\midrule
ALBEF C-A maps                  & 9.4 & 3.7 \\ 
ALBEF ViT maps                  & 18.4 & 9.0 \\ 
\textbf{DINO ViT maps}          & \textbf{21.1} & \textbf{10.5} \\ 
\bottomrule
\end{tabular}
\label{tab:coco_experiments_ablation}
\end{minipage}%
\begin{minipage}[t]{0.48\linewidth}\centering
\caption{Ablation experiments on VOC2007.}
\begin{tabular}{llcc}
\toprule
Method & $\text{mAP}_{50}$ & $\text{mAP}_{50:95}$ \\
\midrule
ALBEF C-A maps                  &   9.2 &  3.3 \\ 
\textbf{ALBEF ViT maps}         &  \textbf{42.9} & \textbf{20.8} \\ 
DINO ViT maps                   &  40.9 & 18.0 \\ 
\bottomrule
\end{tabular}
\label{tab:pascal_experiments_ablation}
\end{minipage}
\end{table}

\section{Conclusion}
\label{sec:Conclusion}

In this paper, we present a two-stage method to locate and label objects by leveraging image-caption pairs without additional supervision. We demonstrate the effectiveness of our approach by performing two tasks in a weakly supervised setting: phrase grounding and object detection. We have performed extensive experiments on Flickr30k Entities, MS COCO and PASCAL VOC2007 achieving state-of-the-art results without the need for supervised object proposal algorithms or text classifiers to process captions. Despite the remarkable performance of our approach, we acknowledge some limitations. Our approach produces a single bounding box per object mentioned in the caption. An interesting direction for further investigation is to extend our method to produce multiple bounding boxes for words representing more than one object instance in the image (\eg "people", "group of animals", \etc). This is particularly challenging, especially when object instances are overlapping in the image. Also, our approach does not generate bounding boxes for objects present in the image but not mentioned in the caption (or due to spelling mistakes). We believe that an important direction for future work is to extend our approach to explicitly take into account missing annotations. Improved performance could be achieved using a more sophisticated pseudo-labeling framework~\cite{xu2019missing,wang2021comining,li2022siod}.

\begin{ack}
This work was conducted as part of the MINDS project of IRT Saint Exupéry. We would like to thank Michelle Aubrun, Ahmad Berjaoui, David Bertoin and Franck Mamalet for useful feedback and suggestions and Jérôme Mathieu for invaluable technical support.
\end{ack}

{\small
\bibliographystyle{plainnat}
\bibliography{egbib}

\begin{thebibliography}{58}
\providecommand{\natexlab}[1]{#1}
\providecommand{\url}[1]{\texttt{#1}}
\expandafter\ifx\csname urlstyle\endcsname\relax
  \providecommand{\doi}[1]{doi: #1}\else
  \providecommand{\doi}{doi: \begingroup \urlstyle{rm}\Url}\fi

\bibitem[Bearman et~al.(2016)Bearman, Russakovsky, Ferrari, and
  Fei-Fei]{bearman2016s}
Amy Bearman, Olga Russakovsky, Vittorio Ferrari, and Li~Fei-Fei.
\newblock What’s the point: Semantic segmentation with point supervision.
\newblock In \emph{Proceedings of the IEEE European Conference on Computer
  Vision}, pages 549--565, 2016.

\bibitem[Bilen and Vedaldi(2016)]{WSDDN}
Hakan Bilen and Andrea Vedaldi.
\newblock Weakly supervised deep detection networks.
\newblock In \emph{Proceedings of the IEEE/CVF Conference on Computer Vision
  and Pattern Recognition}, pages 2846--2854, 2016.

\bibitem[Caron et~al.(2021)Caron, Touvron, Misra, J{\'e}gou, Mairal,
  Bojanowski, and Joulin]{DINO}
Mathilde Caron, Hugo Touvron, Ishan Misra, Herv{\'e} J{\'e}gou, Julien Mairal,
  Piotr Bojanowski, and Armand Joulin.
\newblock Emerging properties in self-supervised vision transformers.
\newblock In \emph{Proceedings of the IEEE/CVF International Conference on
  Computer Vision}, pages 9650--9660, 2021.

\bibitem[Chen et~al.(2020)Chen, Li, Yu, Kholy, Ahmed, Gan, Cheng, and
  Liu]{UNITER}
Yen-Chun Chen, Linjie Li, Licheng Yu, Ahmed~El Kholy, Faisal Ahmed, Zhe Gan,
  Yu~Cheng, and Jingjing Liu.
\newblock Uniter: Universal image-text representation learning.
\newblock In \emph{Proceedings of the IEEE European Conference on Computer
  Vision}, 2020.

\bibitem[Diba et~al.(2017)Diba, Sharma, Pazandeh, Pirsiavash, and
  Van~Gool]{WSCCN}
Ali Diba, Vivek Sharma, Ali Pazandeh, Hamed Pirsiavash, and Luc Van~Gool.
\newblock Weakly supervised cascaded convolutional networks.
\newblock In \emph{Proceedings of the IEEE/CVF Conference on Computer Vision
  and Pattern Recognition}, pages 914--922, 2017.

\bibitem[Dosovitskiy et~al.(2021)Dosovitskiy, Beyer, Kolesnikov, Weissenborn,
  Zhai, Unterthiner, Dehghani, Minderer, Heigold, Gelly, Uszkoreit, and
  Houlsby]{dosovitskiy2021an}
Alexey Dosovitskiy, Lucas Beyer, Alexander Kolesnikov, Dirk Weissenborn,
  Xiaohua Zhai, Thomas Unterthiner, Mostafa Dehghani, Matthias Minderer, Georg
  Heigold, Sylvain Gelly, Jakob Uszkoreit, and Neil Houlsby.
\newblock An image is worth 16x16 words: Transformers for image recognition at
  scale.
\newblock In \emph{Proceedings of the International Conference on Learning
  Representations}, 2021.

\bibitem[Everingham et~al.(2009)Everingham, Van~Gool, Williams, Winn, and
  Zisserman]{PASCALVOC}
Mark Everingham, Luc Van~Gool, Christopher~KI Williams, John Winn, and Andrew
  Zisserman.
\newblock The pascal visual object classes (voc) challenge.
\newblock In \emph{Proceedings of the International Journal of Computer
  Vision}, pages 303--308, 2009.

\bibitem[Gan et~al.(2020)Gan, Chen, Li, Zhu, Cheng, and Liu]{VILLA}
Zhe Gan, Yen-Chun Chen, Linjie Li, Chen Zhu, Yu~Cheng, and Jingjing Liu.
\newblock Large-scale adversarial training for vision-and-language
  representation learning.
\newblock In \emph{Advances in Neural Information Processing Systems}, 2020.

\bibitem[Gao et~al.(2022)Gao, Xing, Niebles, Li, Xu, Liu, and
  Xiong]{gao2022open}
Mingfei Gao, Chen Xing, Juan~Carlos Niebles, Junnan Li, Ran Xu, Wenhao Liu, and
  Caiming Xiong.
\newblock Open vocabulary object detection with pseudo bounding-box labels.
\newblock In \emph{Proceedings of the IEEE European Conference on Computer
  Vision}, pages 266--282, 2022.

\bibitem[Gao et~al.(2019)Gao, Liu, Guo, Ye, Wan, You, and Fan]{gao2019c}
Yan Gao, Boxiao Liu, Nan Guo, Xiaochun Ye, Fang Wan, Haihang You, and Dongrui
  Fan.
\newblock C-midn: Coupled multiple instance detection network with segmentation
  guidance for weakly supervised object detection.
\newblock In \emph{Proceedings of the IEEE/CVF International Conference on
  Computer Vision}, pages 9834--9843, 2019.

\bibitem[Gupta et~al.(2020)Gupta, Vahdat, Chechik, Yang, Kautz, and
  Hoiem]{gupta2020contrastive}
Tanmay Gupta, Arash Vahdat, Gal Chechik, Xiaodong Yang, Jan Kautz, and Derek
  Hoiem.
\newblock Contrastive learning for weakly supervised phrase grounding.
\newblock In \emph{Proceedings of the IEEE European Conference on Computer
  Vision}, pages 752--768, 2020.

\bibitem[He et~al.(2017)He, Gkioxari, Doll{\'a}r, and Girshick]{he2017mask}
Kaiming He, Georgia Gkioxari, Piotr Doll{\'a}r, and Ross Girshick.
\newblock Mask r-cnn.
\newblock In \emph{Proceedings of the IEEE international conference on computer
  vision}, pages 2961--2969, 2017.

\bibitem[Huang et~al.(2020)Huang, Zou, Kumar, and
  Huang]{huang2020comprehensive}
Zeyi Huang, Yang Zou, BVK Kumar, and Dong Huang.
\newblock Comprehensive attention self-distillation for weakly-supervised
  object detection.
\newblock In \emph{Advances in Neural Information Processing Systems}, pages
  16797--16807, 2020.

\bibitem[Jia et~al.(2021)Jia, Yang, Xia, Chen, Parekh, Pham, Le, Sung, Li, and
  Duerig]{ALIGN}
Chao Jia, Yinfei Yang, Ye~Xia, Yi-Ting Chen, Zarana Parekh, Hieu Pham, Quoc Le,
  Yun-Hsuan Sung, Zhen Li, and Tom Duerig.
\newblock Scaling up visual and vision-language representation learning with
  noisy text supervision.
\newblock In \emph{Proceedings of the International Conference on Machine
  Learning}, pages 4904--4916, 2021.

\bibitem[Jocher et~al.(2022)Jocher, Chaurasia, Stoken, Borovec, NanoCode012,
  Kwon, Michael, TaoXie, Fang, imyhxy, Lorna, Yifu, Wong, V, Montes, Wang,
  Fati, Nadar, Laughing, UnglvKitDe, Sonck, tkianai, yxNONG, Skalski, Hogan,
  Nair, Strobel, and Jain]{YOLOv5}
Glenn Jocher, Ayush Chaurasia, Alex Stoken, Jirka Borovec, NanoCode012, Yonghye
  Kwon, Kalen Michael, TaoXie, Jiacong Fang, imyhxy, Lorna, Zeng Yifu, Colin
  Wong, Abhiram V, Diego Montes, Zhiqiang Wang, Cristi Fati, Jebastin Nadar,
  Laughing, UnglvKitDe, Victor Sonck, tkianai, yxNONG, Piotr Skalski, Adam
  Hogan, Dhruv Nair, Max Strobel, and Mrinal Jain.
\newblock {ultralytics/yolov5: v7.0 - YOLOv5 SOTA Realtime Instance
  Segmentation}, November 2022.
\newblock URL \url{https://doi.org/10.5281/zenodo.7347926}.

\bibitem[Kim et~al.(2021)Kim, Son, and Kim]{VILT}
Wonjae Kim, Bokyung Son, and Ildoo Kim.
\newblock Vilt: Vision-and-language transformer without convolution or region
  supervision.
\newblock In \emph{Proceedings of the International Conference on Machine
  Learning}, pages 5583--5594. PMLR, 2021.

\bibitem[Krishna et~al.(2017)Krishna, Zhu, Groth, Johnson, Hata, Kravitz, Chen,
  Kalantidis, Li, Shamma, et~al.]{krishna2017visual}
Ranjay Krishna, Yuke Zhu, Oliver Groth, Justin Johnson, Kenji Hata, Joshua
  Kravitz, Stephanie Chen, Yannis Kalantidis, Li-Jia Li, David~A Shamma, et~al.
\newblock Visual genome: Connecting language and vision using crowdsourced
  dense image annotations.
\newblock \emph{International Journal of Computer Vision}, pages 32--73, 2017.

\bibitem[Li et~al.(2022{\natexlab{a}})Li, Pan, Yan, Tang, and
  Zheng]{li2022siod}
Hanjun Li, Xingjia Pan, Ke~Yan, Fan Tang, and Wei-Shi Zheng.
\newblock Siod: single instance annotated per category per image for object
  detection.
\newblock In \emph{Proceedings of the IEEE/CVF Conference on Computer Vision
  and Pattern Recognition}, pages 14197--14206, 2022{\natexlab{a}}.

\bibitem[Li et~al.(2021)Li, Selvaraju, Gotmare, Joty, Xiong, and Hoi]{ALBEF}
Junnan Li, Ramprasaath~R. Selvaraju, Akhilesh~Deepak Gotmare, Shafiq Joty,
  Caiming Xiong, and Steven Hoi.
\newblock Align before fuse: Vision and language representation learning with
  momentum distillation.
\newblock In A.~Beygelzimer, Y.~Dauphin, P.~Liang, and J.~Wortman Vaughan,
  editors, \emph{Advances in Neural Information Processing Systems}, 2021.
\newblock URL \url{https://openreview.net/forum?id=OJLaKwiXSbx}.

\bibitem[Li et~al.(2019)Li, Yatskar, Yin, Hsieh, and Chang]{VisualBERT}
Liunian~Harold Li, Mark Yatskar, Da~Yin, Cho-Jui Hsieh, and Kai-Wei Chang.
\newblock Visualbert: A simple and performant baseline for vision and language.
\newblock \emph{arXiv preprint arXiv:1908.03557}, 2019.

\bibitem[Li et~al.(2022{\natexlab{b}})Li, Zhang, Zhang, Yang, Li, Zhong, Wang,
  Yuan, Zhang, Hwang, et~al.]{GLIP}
Liunian~Harold Li, Pengchuan Zhang, Haotian Zhang, Jianwei Yang, Chunyuan Li,
  Yiwu Zhong, Lijuan Wang, Lu~Yuan, Lei Zhang, Jenq-Neng Hwang, et~al.
\newblock Grounded language-image pre-training.
\newblock In \emph{Proceedings of the IEEE/CVF Conference on Computer Vision
  and Pattern Recognition}, pages 10965--10975, 2022{\natexlab{b}}.

\bibitem[Li et~al.(2020)Li, Yin, Li, Hu, Zhang, Zhang, Wang, Hu, Dong, Wei,
  Choi, and Gao]{OSCAR}
Xiujun Li, Xi~Yin, Chunyuan Li, Xiaowei Hu, Pengchuan Zhang, Lei Zhang, Lijuan
  Wang, Houdong Hu, Li~Dong, Furu Wei, Yejin Choi, and Jianfeng Gao.
\newblock Oscar: Object-semantics aligned pre-training for vision-language
  tasks.
\newblock In \emph{Proceedings of the IEEE European Conference on Computer
  Vision}, 2020.

\bibitem[Lin et~al.(2014)Lin, Maire, Belongie, Hays, Perona, Ramanan,
  Doll{\'a}r, and Zitnick]{MSCOCO}
Tsung-Yi Lin, Michael Maire, Serge Belongie, James Hays, Pietro Perona, Deva
  Ramanan, Piotr Doll{\'a}r, and C~Lawrence Zitnick.
\newblock Microsoft coco: Common objects in context.
\newblock In \emph{Proceedings of the IEEE European Conference on Computer
  Vision}, pages 740--755. Springer, 2014.

\bibitem[Liu et~al.(2021)Liu, Ma, He, Kuo, Chen, Zhang, Wu, Kira, and
  Vajda]{liu2021unbiased}
Yen-Cheng Liu, Chih-Yao Ma, Zijian He, Chia-Wen Kuo, Kan Chen, Peizhao Zhang,
  Bichen Wu, Zsolt Kira, and Peter Vajda.
\newblock Unbiased teacher for semi-supervised object detection.
\newblock In \emph{Proceedings of the International Conference on Learning
  Representations}, 2021.
\newblock URL \url{https://openreview.net/forum?id=MJIve1zgR_}.

\bibitem[Liu et~al.(2022)Liu, Ma, and Kira]{liu2022unbiased}
Yen-Cheng Liu, Chih-Yao Ma, and Zsolt Kira.
\newblock Unbiased teacher v2: Semi-supervised object detection for anchor-free
  and anchor-based detectors.
\newblock In \emph{Proceedings of the IEEE/CVF Conference on Computer Vision
  and Pattern Recognition}, pages 9819--9828, 2022.

\bibitem[Lu et~al.(2019)Lu, Batra, Parikh, and Lee]{ViLBERT}
Jiasen Lu, Dhruv Batra, Devi Parikh, and Stefan Lee.
\newblock Vilbert: Pretraining task-agnostic visiolinguistic representations
  for vision-and-language tasks.
\newblock In \emph{Advances in Neural Information Processing Systems}, pages
  13--23, 2019.

\bibitem[Lu et~al.(2020)Lu, Goswami, Rohrbach, Parikh, and Lee]{ViLBERT2}
Jiasen Lu, Vedanuj Goswami, Marcus Rohrbach, Devi Parikh, and Stefan Lee.
\newblock 12-in-1: Multi-task vision and language representation learning.
\newblock In \emph{Proceedings of the IEEE/CVF Conference on Computer Vision
  and Pattern Recognition}, June 2020.

\bibitem[Maron and Lozano-P{\'e}rez(1997)]{maron1997framework}
Oded Maron and Tom{\'a}s Lozano-P{\'e}rez.
\newblock A framework for multiple-instance learning.
\newblock In \emph{Advances in Neural Information Processing Systems}, 1997.

\bibitem[Plummer et~al.(2015)Plummer, Wang, Cervantes, Caicedo, Hockenmaier,
  and Lazebnik]{Flickr30kEntities}
Bryan~A Plummer, Liwei Wang, Chris~M Cervantes, Juan~C Caicedo, Julia
  Hockenmaier, and Svetlana Lazebnik.
\newblock Flickr30k entities: Collecting region-to-phrase correspondences for
  richer image-to-sentence models.
\newblock In \emph{Proceedings of the IEEE/CVF International Conference on
  Computer Vision}, pages 2641--2649, 2015.

\bibitem[Radford et~al.(2021)Radford, Kim, Hallacy, Ramesh, Goh, Agarwal,
  Sastry, Askell, Mishkin, Clark, Krueger, and Sutskever]{CLIP}
Alec Radford, Jong~Wook Kim, Chris Hallacy, Aditya Ramesh, Gabriel Goh,
  Sandhini Agarwal, Girish Sastry, Amanda Askell, Pamela Mishkin, Jack Clark,
  Gretchen Krueger, and Ilya Sutskever.
\newblock Learning transferable visual models from natural language
  supervision.
\newblock \emph{arXiv preprint arXiv:2103.00020}, 2021.

\bibitem[Ren et~al.(2015)Ren, He, Girshick, and Sun]{ren2015faster}
Shaoqing Ren, Kaiming He, Ross Girshick, and Jian Sun.
\newblock Faster r-cnn: Towards real-time object detection with region proposal
  networks.
\newblock In \emph{Advances in Neural Information Processing Systems}, 2015.

\bibitem[Ren et~al.(2020)Ren, Yu, Yang, Liu, Lee, Schwing, and
  Kautz]{ren2020instance}
Zhongzheng Ren, Zhiding Yu, Xiaodong Yang, Ming-Yu Liu, Yong~Jae Lee,
  Alexander~G Schwing, and Jan Kautz.
\newblock Instance-aware, context-focused, and memory-efficient weakly
  supervised object detection.
\newblock In \emph{Proceedings of the IEEE/CVF Conference on Computer Vision
  and Pattern Recognition}, pages 10598--10607, 2020.

\bibitem[Selvaraju et~al.(2017)Selvaraju, Cogswell, Das, Vedantam, Parikh, and
  Batra]{selvaraju2017grad}
Ramprasaath~R Selvaraju, Michael Cogswell, Abhishek Das, Ramakrishna Vedantam,
  Devi Parikh, and Dhruv Batra.
\newblock Grad-cam: Visual explanations from deep networks via gradient-based
  localization.
\newblock In \emph{Proceedings of the IEEE European Conference on Computer
  Vision}, pages 618--626, 2017.

\bibitem[Sharma et~al.(2018)Sharma, Ding, Goodman, and
  Soricut]{sharma2018conceptual}
Piyush Sharma, Nan Ding, Sebastian Goodman, and Radu Soricut.
\newblock Conceptual captions: A cleaned, hypernymed, image alt-text dataset
  for automatic image captioning.
\newblock In \emph{Proceedings of the 56th Annual Meeting of the Association
  for Computational Linguistics}, pages 2556--2565, 2018.

\bibitem[Shi et~al.(2022)Shi, Hayat, Wu, and Cai]{shi2022proposalclip}
Hengcan Shi, Munawar Hayat, Yicheng Wu, and Jianfei Cai.
\newblock Proposalclip: unsupervised open-category object proposal generation
  via exploiting clip cues.
\newblock In \emph{Proceedings of the IEEE/CVF Conference on Computer Vision
  and Pattern Recognition}, pages 9611--9620, 2022.

\bibitem[Sim\'eoni et~al.(2021)Sim\'eoni, Puy, Vo, Roburin, Gidaris, Bursuc,
  P\'erez, Marlet, and Ponce]{LOST}
Oriane Sim\'eoni, Gilles Puy, Huy~V. Vo, Simon Roburin, Spyros Gidaris, Andrei
  Bursuc, Patrick P\'erez, Renaud Marlet, and Jean Ponce.
\newblock Localizing objects with self-supervised transformers and no labels.
\newblock In \emph{Proceedings of the British Machine Vision Conference}, 2021.

\bibitem[Sohn et~al.(2020)Sohn, Zhang, Li, Zhang, Lee, and Pfister]{STAC}
Kihyuk Sohn, Zizhao Zhang, Chun-Liang Li, Han Zhang, Chen-Yu Lee, and Tomas
  Pfister.
\newblock A simple semi-supervised learning framework for object detection.
\newblock In \emph{arXiv:2005.04757}, 2020.

\bibitem[Su et~al.(2020)Su, Zhu, Cao, Li, Lu, Wei, and Dai]{VLBERT}
Weijie Su, Xizhou Zhu, Yue Cao, Bin Li, Lewei Lu, Furu Wei, and Jifeng Dai.
\newblock Vl-bert: Pre-training of generic visual-linguistic representations.
\newblock In \emph{Proceedings of the International Conference on Learning
  Representations}, 2020.
\newblock URL \url{https://openreview.net/forum?id=SygXPaEYvH}.

\bibitem[Tan and Bansal(2019)]{LXMERT}
Hao Tan and Mohit Bansal.
\newblock Lxmert: Learning cross-modality encoder representations from
  transformers.
\newblock In \emph{Proceedings of the Conference on Empirical Methods in
  Natural Language Processing}, 2019.

\bibitem[Tang et~al.(2018)Tang, Wang, Bai, Shen, Bai, Liu, and
  Yuille]{tang2018pcl}
Peng Tang, Xinggang Wang, Song Bai, Wei Shen, Xiang Bai, Wenyu Liu, and Alan
  Yuille.
\newblock Pcl: Proposal cluster learning for weakly supervised object
  detection.
\newblock In \emph{IEEE Transactions on Pattern Analysis and Machine
  Intelligence}, pages 176--191, 2018.

\bibitem[Uijlings et~al.(2013)Uijlings, Van De~Sande, Gevers, and
  Smeulders]{uijlings2013selective}
Jasper~RR Uijlings, Koen~EA Van De~Sande, Theo Gevers, and Arnold~WM Smeulders.
\newblock Selective search for object recognition.
\newblock In \emph{Proceedings of the International Journal of Computer
  Vision}, pages 154--171, 2013.

\bibitem[Vo et~al.(2022)Vo, Sim{\'e}oni, Gidaris, Bursuc, P{\'e}rez, and
  Ponce]{vo2022active}
Huy~V Vo, Oriane Sim{\'e}oni, Spyros Gidaris, Andrei Bursuc, Patrick P{\'e}rez,
  and Jean Ponce.
\newblock Active learning strategies for weakly-supervised object detection.
\newblock In \emph{Proceedings of the IEEE European Conference on Computer
  Vision}, pages 211--230, 2022.

\bibitem[Wang et~al.(2018)Wang, Yan, Zhang, Zhang, and Lin]{wang2018towards}
Keze Wang, Xiaopeng Yan, Dongyu Zhang, Lei Zhang, and Liang Lin.
\newblock Towards human-machine cooperation: Self-supervised sample mining for
  object detection.
\newblock In \emph{Proceedings of the IEEE/CVF Conference on Computer Vision
  and Pattern Recognition}, pages 1605--1613, 2018.

\bibitem[Wang et~al.(2021)Wang, Yang, Cao, and Zhang]{wang2021comining}
Tiancai Wang, Tong Yang, Jiale Cao, and Xiangyu Zhang.
\newblock Co-mining: Self-supervised learning for sparsely annotated object
  detection.
\newblock In \emph{Proceedings of the AAAI Conference on Artificial
  Intelligence}, 2021.

\bibitem[Wang et~al.(2022)Wang, Shen, Hu, Yuan, Crowley, and
  Vaufreydaz]{wang2022self}
Yangtao Wang, Xi~Shen, Shell~Xu Hu, Yuan Yuan, James~L Crowley, and Dominique
  Vaufreydaz.
\newblock Self-supervised transformers for unsupervised object discovery using
  normalized cut.
\newblock In \emph{Proceedings of the IEEE/CVF Conference on Computer Vision
  and Pattern Recognition}, pages 14543--14553, 2022.

\bibitem[Xu et~al.(2019)Xu, Bai, Ghanem, Liu, Gao, Guo, Ye, Wan, You, Fan,
  et~al.]{xu2019missing}
Mengmeng Xu, Yancheng Bai, Bernard Ghanem, Boxiao Liu, Yan Gao, N~Guo, X~Ye,
  F~Wan, H~You, D~Fan, et~al.
\newblock Missing labels in object detection.
\newblock In \emph{Proceedings of the IEEE/CVF Conference on Computer Vision
  and Pattern Recognition Workshops}, page~5, 2019.

\bibitem[Yang et~al.(2017)Yang, Zhang, Rehman, and Huang]{yang2017image}
Zhongliang Yang, Yu-Jin Zhang, Sadaqat~ur Rehman, and Yongfeng Huang.
\newblock Image captioning with object detection and localization.
\newblock In \emph{Proceedings of the International Conference on Image and
  Graphics}, pages 109--118, 2017.

\bibitem[Ye et~al.(2019)Ye, Zhang, Kovashka, Li, Qin, and Berent]{CAP2DET}
Keren Ye, Mingda Zhang, Adriana Kovashka, Wei Li, Danfeng Qin, and Jesse
  Berent.
\newblock Cap2det: Learning to amplify weak caption supervision for object
  detection.
\newblock In \emph{Proceedings of the IEEE/CVF International Conference on
  Computer Vision}, pages 9686--9695, 2019.

\bibitem[Young et~al.(2014)Young, Lai, Hodosh, and Hockenmaier]{young2014image}
Peter Young, Alice Lai, Micah Hodosh, and Julia Hockenmaier.
\newblock From image descriptions to visual denotations: New similarity metrics
  for semantic inference over event descriptions.
\newblock \emph{Transactions of the Association for Computational Linguistics},
  pages 67--78, 2014.

\bibitem[Yu et~al.(2016)Yu, Poirson, Yang, Berg, and Berg]{yu2016modeling}
Licheng Yu, Patrick Poirson, Shan Yang, Alexander~C Berg, and Tamara~L Berg.
\newblock Modeling context in referring expressions.
\newblock In \emph{Proceedings of the IEEE European Conference on Computer
  Vision}, pages 69--85. Springer, 2016.

\bibitem[Yu et~al.(2018)Yu, Lin, Shen, Yang, Lu, Bansal, and
  Berg]{yu2018mattnet}
Licheng Yu, Zhe Lin, Xiaohui Shen, Jimei Yang, Xin Lu, Mohit Bansal, and
  Tamara~L Berg.
\newblock Mattnet: Modular attention network for referring expression
  comprehension.
\newblock In \emph{Proceedings of the IEEE/CVF Conference on Computer Vision
  and Pattern Recognition}, pages 1307--1315, 2018.

\bibitem[Zareian et~al.(2021)Zareian, Rosa, Hu, and Chang]{zareian2021open}
Alireza Zareian, Kevin~Dela Rosa, Derek~Hao Hu, and Shih-Fu Chang.
\newblock Open-vocabulary object detection using captions.
\newblock In \emph{Proceedings of the IEEE/CVF Conference on Computer Vision
  and Pattern Recognition}, pages 14393--14402, 2021.

\bibitem[Zeng et~al.(2019)Zeng, Liu, Fu, Chao, and Zhang]{zeng2019wsod2}
Zhaoyang Zeng, Bei Liu, Jianlong Fu, Hongyang Chao, and Lei Zhang.
\newblock Wsod2: Learning bottom-up and top-down objectness distillation for
  weakly-supervised object detection.
\newblock In \emph{Proceedings of the IEEE/CVF International Conference on
  Computer Vision}, pages 8292--8300, 2019.

\bibitem[Zhang et~al.(2020)Zhang, Chen, Shen, Hao, Zhu, and
  Savvides]{zhang2020solving}
Han Zhang, Fangyi Chen, Zhiqiang Shen, Qiqi Hao, Chenchen Zhu, and Marios
  Savvides.
\newblock Solving missing-annotation object detection with background
  recalibration loss.
\newblock In \emph{IEEE International Conference on Acoustics, Speech and
  Signal Processing}, pages 1888--1892, 2020.

\bibitem[Zhang et~al.(2016)Zhang, Shang, Yang, Xu, Luan, and
  Chua]{zhang2016online}
Hanwang Zhang, Xindi Shang, Wenzhuo Yang, Huan Xu, Huanbo Luan, and Tat-Seng
  Chua.
\newblock Online collaborative learning for open-vocabulary visual classifiers.
\newblock In \emph{Proceedings of the IEEE/CVF Conference on Computer Vision
  and Pattern Recognition}, pages 2809--2817, 2016.

\bibitem[Zhang et~al.(2018)Zhang, Bai, Ding, Li, and Ghanem]{W2F}
Yongqiang Zhang, Yancheng Bai, Mingli Ding, Yongqiang Li, and Bernard Ghanem.
\newblock W2f: A weakly-supervised to fully-supervised framework for object
  detection.
\newblock In \emph{Proceedings of the IEEE/CVF Conference on Computer Vision
  and Pattern Recognition}, pages 928--936, 2018.

\bibitem[Zhong et~al.(2022)Zhong, Yang, Zhang, Li, Codella, Li, Zhou, Dai,
  Yuan, Li, et~al.]{zhong2022regionclip}
Yiwu Zhong, Jianwei Yang, Pengchuan Zhang, Chunyuan Li, Noel Codella,
  Liunian~Harold Li, Luowei Zhou, Xiyang Dai, Lu~Yuan, Yin Li, et~al.
\newblock Regionclip: Region-based language-image pretraining.
\newblock In \emph{Proceedings of the IEEE/CVF Conference on Computer Vision
  and Pattern Recognition}, pages 16793--16803, 2022.

\bibitem[Zitnick and Doll{\'a}r(2014)]{zitnick2014edge}
C~Lawrence Zitnick and Piotr Doll{\'a}r.
\newblock Edge boxes: Locating object proposals from edges.
\newblock In \emph{Proceedings of the IEEE European Conference on Computer
  Vision}, pages 391--405, 2014.

\end{thebibliography}
}


\end{document}